\documentclass{article}
\usepackage{arxiv}
\usepackage[utf8]{inputenc}
\usepackage[T1]{fontenc}
\usepackage{hyperref}
\usepackage{url}
\usepackage{booktabs}
\usepackage{amsfonts}
\usepackage{nicefrac}
\usepackage{microtype}
\usepackage{lipsum}
\usepackage{graphicx}
\graphicspath{ {./images/} }

\usepackage{times}
\usepackage{latexsym}
\usepackage{amsmath}
\usepackage{amssymb}
\usepackage{amsfonts}
\usepackage{booktabs}
\usepackage{threeparttable}
\usepackage{url}
\usepackage{graphicx}
\usepackage{epsfig}
\usepackage{cite}
\usepackage{multirow}
\usepackage{comment}
\usepackage{subfigure}

\title{Multichannel CNN with Attention for Text Classification}

\author{
 Zhenyu~Liu \\
  Department of Science and Technology Teaching\\
  China University of Political Science and Law\\
  Beijing 100088\\
  \texttt{lzhy@cupl.edu.cn} \\
   \And
 Haiwei~Huang \\
  School of Coumputer and Science Technology\\
  University of Science and Technique of China\\
  Hefei, Anhui 230027\\
  \texttt{hwhuang@mail.ustc.edu.cn} \\
   \And
 Chaohong~Lu \\
  School of Coumputer and Science Technology\\
  University of Science and Technique of China\\
  Hefei, Anhui 230027 \\
  \texttt{lchwhut@mail.ustc.edu.cn} \\
  \And
 Shengfei~Lyu \\
  School of Coumputer and Science Technology\\
  University of Science and Technique of China\\
  Hefei, Anhui 230027 \\
  \texttt{saintfe@mail.ustc.edu.cn} \\
}

\begin{document}
\maketitle
\begin{abstract}
Recent years, the approaches based on neural networks have shown remarkable potential for sentence modeling. There are two main neural network structures: recurrent neural network (RNN) and convolution neural network (CNN). RNN can capture long term dependencies and store the semantics of the previous information in a fixed-sized vector. However, RNN is a biased model and its ability to extract global semantics is restricted by the fixed-sized vector. Alternatively, CNN is able to capture n-gram features of texts by utilizing convolutional filters. But the width of convolutional filters restricts its performance. In order to combine the strengths of the two kinds of networks and alleviate their shortcomings, this paper proposes Attention-based Multichannel Convolutional Neural Network (AMCNN) for text classification. AMCNN utilizes a bi-directional long short-term memory to encode the history and future information of words into high dimensional representations, so that the information of  both the front and back of the sentence can be fully expressed. Then the scalar attention and vectorial attention are applied to obtain multichannel representations. The scalar attention can calculate the word-level importance and the vectorial attention can calculate the feature-level importance. In the classification task, AMCNN uses a CNN structure to cpture word relations on the representations generated by the scalar and vectorial attention mechanism instead of calculating the weighted sums. It can effectively extract the n-gram features of the text. The experimental results on the benchmark datasets demonstrate that AMCNN achieves better performance than state-of-the-art methods. In addition, the visualization results verify the semantic richness of multichannel representations.
\end{abstract}


\section{Introduction}

Natural Language Processing (NLP) is an important field in
Artificial Intelligence (AI) and it includes many interesting
research topics such as text classification\cite{Jiang2018LatentTT}.
Text classification aims to categorize texts into different classes.
Traditional methods tend to apply the bag-of-words (BOW) model to
represent texts as unordered sets and input them to classification
algorithms such as support vector machines (SVM)
\cite{vapnik1998statistical} and its probabilistic version, e.g.
probabilistic classification vector machines
\cite{chen-2009-pcvm,chen-2014-epcvm}, and its multi-objective version \cite{lyu2019multiclass} 
and large-scale version \cite{jiang2017scalable}. However, text features
obtained by the BOW model fail to capture the semantics of texts due
to the loss of word order information. It also suffers from the
problems of high dimension and data sparsity. To solve these
problems, word embedding is proposed as the distributed
representations of words. This dense representation outperforms
traditional methods such as one-hot representation since it can
alleviate data sparsity and make semantically similar words close
with cosine or Euclidean distance. It becomes the mainstream for NLP
tasks to learn distributed representations of words through neural
language models \cite{bengio2003neural, le2014distributed} and
perform a combination of learned word representation with attention
mechanism \cite{luong2015effective}. Some researches show that the
character-level representations also work well on the specific tasks
\cite{zhang2015character,xu-etal-2016-improve} and can achieve
competitive results.

Recurrent Neural Network (RNN) is a popular neural network
architecture for dealing with NLP tasks owing to its capability of
handing sequences of any length \cite{tang2015document}. This
architecture sequentially processes word by word and generates a
hidden state at each time step to represent all previous words. RNN
has the ability to capture the long-term dependencies of a text but
it might suffer from gradient exploding or gradient vanishing
problem. To overcome this shortcoming, long short-term memory
network (LSTM) \cite{hochreiter1997long,zhao2020condition} and other variants such as
gated recurrent unit (GRU) \cite{cho2014learning}, simple recurrent
unit (SRU) \cite{lei2017training} were proposed for better
remembering and memory accesses. When RNN is used to process a long
sentence, the recent information is more dominant than earlier
information that may be the real dominant part of the sentence.
However, the most important information could appear anywhere in a
sentence rather than at the end. Therefore, some researchers propose
to average the hidden states of all time steps in RNN and assign the
same weight to all hidden states by default.

In order to better select the input elements needed for downstream
tasks, attention mechanism is introduced to assign different importance
weights to all outputs of RNN. Intuitively, the attention mechanism
enables RNN to maintain a variable-length memory and merge the
outputs according to their importance weights. It has been proven to
be effective in NLP tasks such as neural machine translation
\cite{luong2015effective,cho2014learning}. However, the attention
mechanism fails to capture the relations of words and has the
disadvantage of losing the word order information, which is
important to downstream tasks. Taking the sentences ``\textit{Tina
likes Bob.}" and ``\textit{Bob likes Tina.}" as examples, the
weighted sum of their hidden states encoded by RNN are nearly
similar. But the two sentences mean exactly the opposite
situations.

Convolution Neural Network (CNN) is another popular architecture for
dealing with NLP tasks. CNN performs excellently in extracting
n-gram features by utilizing convolutional filters to capture local
correlations in a parallel way. For the above example, 2-gram
features of `` \textit{Tina likes}" and ``\textit{likes Bob}" could
be captured by CNN. These features are more representative of the
original sentence than the weighted sum of hidden states. CNN has
been proposed for tackling NLP tasks and has achieved remarkable
results in sentence modeling \cite{kalchbrenner2014convolutional},
semantic parsing \cite{yih2014semantic}, and text classification
\cite{kim2014convolutional}. But its ability to extract n-gram
features is subject to the width of convolutional filters.

Some researchers introduced hybrid frameworks of CNN and RNN to
combine their advantages. The proposed recurrent convolutional
neural network \cite{lai2015recurrent} applied a recurrent
convolutional structure to capture contextual information as far as
possible. The method employed a max-pooling layer to select key
words for the purpose of introducing less noise compared to
window-based neural networks. As a similar work, convolutional
recurrent neural network\cite{wang2017hybrid} integrated the merits
to extract different aspects of linguistic information from both
convolutional and recurrent neural network structures. However, most
hybrid frameworks treated all words equally and neglected the fact
that different words have different contributions to the semantics
of a text.

To take full advantage of CNN and alleviate its shortcomings, we
propose Attention-based Multichannel Convolutional Neural Network
(AMCNN) that equips RNN and CNN with the help of the scalar
attention and the vectorial attention. The contributions of our work
are listed as follows:

\begin{figure}
   \centering
   \includegraphics[width=0.9\linewidth]{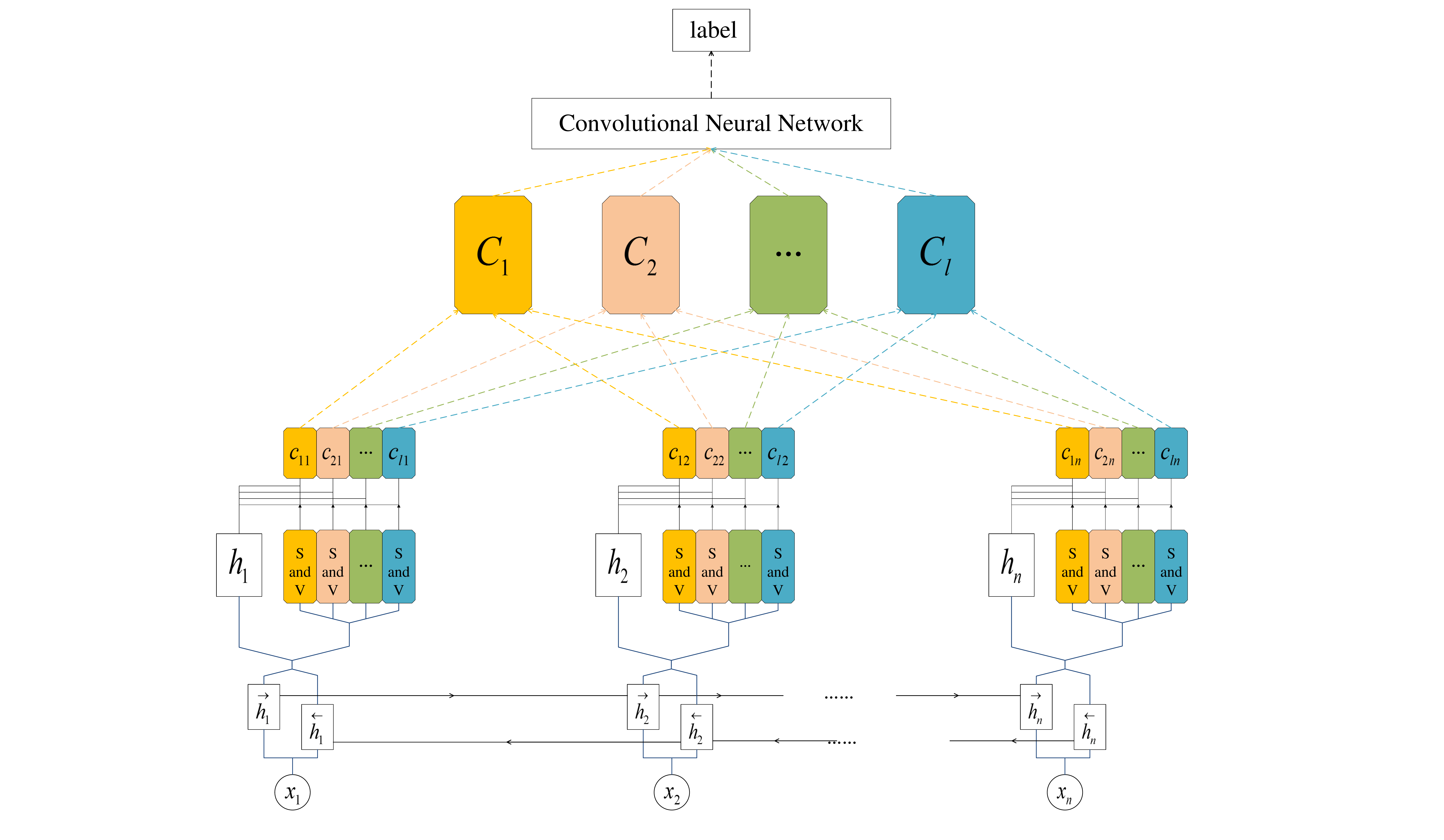}
   \caption{The structure of the Attention-based Multichannel Convolutional Neural Network. S and V denote the scalar attention and the vectorial attention, respectively. Blocks of the same color are merged into one channel}
   \label{fig:model}
\end{figure}

\begin{enumerate}
    \item AMCNN uses a bi-directional LSTM to generate forward hidden states and
    backward hidden states in each time step, and then make them interact non-linearly.
    The combination of the two types of hidden states can make full use of word order information
    and will not lose the information at both ends of long sentences.

    \item Attention mechanisms we proposed could help generate multichannel which is
    regarded as diversification of input information caused by data perturbation.
    The scalar attention and the vectorial attention can calculate the word-level and
    feature-level importance, respectively.
    And the multichannel representations reflect the different contributions of different words
    to the semantics of a text.

    \item AMCNN utilizes CNN to capture word relations by applying the attention mechanism instead of calculating weighted sum. In this way, the ability of CNN to extract n-gram features has also been enhanced.
\end{enumerate}

This paper is organized as follows. Section \ref{sec:Related work}
introduces the related work about CNN and attention mechanisms.
Section \ref{sec:Method} introduces the proposed AMCNN in detail.
And Section \ref{sec:Experiment} introduces datasets, baselines,
experiments, and analysis. Finally, Section \ref{sec:Conclusion}
concludes this paper.

\section{Related work}
\label{sec:Related work}

Most of the previous work has exploited deep learning to deal with
NLP tasks, including learning distributed representations of words,
sentences or documents
\cite{mikolov2013distributed,le2014distributed,kalchbrenner2014convolutional,wang2017hybrid}
and text classification
\cite{zhang2015character,yang2016hierarchical,joulin2016bag,lai2015recurrent},
etc.

A CNN architecture \cite{kim2014convolutional} was proposed with
multiple filters to capture local correlations followed by
max-pooling operation to extract dominant features. This
architecture performs well on text classification with a few
parameters. The case of using character-level CNN was explored for
text classification without word embedding \cite{zhang2015character}
and in this work language was regarded as a kind of signal. Based on
character-level representations, very deep convolutional networks
(VDCNN)\cite{conneau2016very} were applied to text classification
which is up to 29 convolutional layers much larger than 1 layer used
by \cite{kim2014convolutional}. To capture word correlations of
different sizes, a dynamic $k$-max-pooling method, a global pooling
operation over linear sequences, was proposed to keep features
better\cite{kalchbrenner2014convolutional}. Tree-structured
sentences were also explored convolutional
models\cite{mou2015discriminative}. Multichannel variable-size
convolution neural network (MVCNN) \cite{yin2016multichannel}
combined diverse versions of pre-trained word embedding and used
varied-size convolution filters to extract features.

A RNN is often employed to process temporal sequences. Instead of RNN, there are several approaches
for sequences learning, such as echo state network and learning in the model space \cite{li2018symbolic,gong2018sequential,chen2014cognitive}. In the learning in the model space, it transforms the original temporal series to an echo state network (ESN), and calculates the `distance' between ESNs  \cite{ChenTRY14,chen2013model}. Therefore, the distance based learning algorithms could be employed in the 
ESN space \cite{gong2016model}. Chen et al. \cite{chen2015model} investigated the trade-off between the representation and discrimination abilities. Gong et al. proposed the multi-objective version for learning in the model space \cite{gong2018multiobjective}. 

The other popular RNN architecture is able to deal with input
sequences of varied length and capture long-term dependencies. Gated
recurrent neural network (GRU) \cite{chung2014empirical} was
proposed to model sequences. As a similar work, GRU was applied to
model documents\cite{tang2015document}. Their works show that GRU
has the ability to encode relations between sentences in a document.
To improve the performance of GRU on large scale text, hierarchical
attention networks (HAN)\cite{yang2016hierarchical} was proposed.
HAN has a hierarchical structure including word encoder and sentence
encoder with two levels of attention mechanisms.

As an auxiliary way to select inputs, attention mechanism is widely
adopted in deep learning recently due to its flexibility in modeling
dependencies and parallelized calculation. The attention mechanism
was introduced to improve encoder-decoder based neural machine
translation \cite{bahdanau2014neural}. It allows a model to
automatically search for parts of elements that are related to the
target word. As an extension, global attention and local attention
\cite{luong2015effective} were proposed to deal with machine
translation and their alignment visualizations proved the ability to
learn dependencies. In HAN \cite{yang2016hierarchical}, hierarchical
attention was used to generate document-level representations from
word-level representations and sentence-level representations. This
architecture simply sets a trainable context vector as a high-level
representation of a fixed query. This way may be unsuitable because
the same words may count differently in varied contexts. In a recent
work \cite{vaswani2017attention}, the calculation of attention
mechanism was generalized into Q-K-V\footnote{Q-K-V denotes query,
key and value respectively.} form.

\section{Attention-based Multichannel Convolutional Neural Network}
\label{sec:Method}

The architecture of attention-based multichannel convolutional
neural network (AMCNN) is demonstrated in Fig.\ref{fig:model}. It
consists of three parts: bi-directional long short-term memory
(Bi-LSTM), attention layer and convolutional neural network (CNN).
The following sections describe how we utilize Bi-LSTM to generate
the scalar attention and the vectorial attention, and form
multichannel for CNN.

\subsection{Long Short-Term Memory Network}

In many NLP tasks, RNN processes word embedding for texts of
variable length and generates a hidden state $ h_{t} $ in $t$ time
step by recursively transforming the previous hidden state $
h_{t-1}$ and the current input vector $ x_t $.
\begin{equation}
{h_t} = f(W \cdot [{h_{t - 1}},{x_t}] + b),
\end{equation}
where $ W \in {\mathbb{R}^{{l_h} \times \left( {{l_h} + {l_i}} \right)}} $ , $ b \in {\mathbb{R}^{{l_h}}} $ , $ l_h $ and $ l_i $ are dimensions of hidden state and input vector respectively, and $ f\left(  \cdot  \right) $ represents activation function such as $ tanh\left(  \cdot  \right) $. However, standard RNN is not a preferable choice for researchers due to the problem of gradient exploding or vanishing \cite{bengio1994learning}. To address this problem, LSTM was introduced and obtained remarkable performance.

As a kind of variant of RNN, the LSTM architecture has a range of
tandem modules whose parameters are shared. At $t$ time step, the
hidden state $ h_t $ is controlled by the previous hidden state $
h_{t-1} $, input $ x_t $, forget gate $ f_t $, input gate $ i_t $
and output gate $ o_t $. These gates identify the way of updating
the current memory cell $ c_t $ and the current hidden state $ h_t
$. The LSTM transition function can be summarized by the following
equations:
\begin{equation}
\begin{split}
{f_t} &= \sigma ({W_f} \cdot \left[ {{h_{t - 1}},{x_t}} \right] + {b_f}), \\
{i_t} &= \sigma ({W_i} \cdot \left[ {{h_{t - 1}},{x_t}} \right] + {b_i}), \\
{o_t} &= \sigma ({W_o} \cdot \left[ {{h_{t - 1}},{x_t}} \right] + {b_o}), \\
\mathop {{C_t}}\limits^ \sim   &= \tanh ({W_C} \cdot \left[ {{h_{t - 1}},{x_t}} \right] + {b_C}), \\
{C_t} &= {f_t} \odot {C_{t - 1}} + {i_t} \odot \mathop {{C_t}}\limits^ \sim,  \\
{h_t} &= {o_t} \odot \tanh ({C_t}). \\
\end{split}
\end{equation}

Here, $ \sigma $ is the logistic sigmoid function that has the
domain of all real numbers, with return value ranging from 0 to 1. $
tanh $ denotes the hyperbolic tangent function with return value
ranging from -1 to 1. Intuitively, the forget gate $ f_t $ controls
the extent to which the previous cell state $ C_{t-1} $ remains in
the cell. The input gate $ i_t $ controls the extent to which a new
input flows into the cell. The output gate $ o_t $ controls the
extent to which the cell state $ C_t $ is used to compute the
current hidden state $ h_t $. The existence of those gates enables
LSTM to capture long-term dependencies  when dealing with
time-series data.


Though unidirectional LSTM includes an unbounded sentence history in
theory, it is still constrained since the hidden state of each time
step fails to model future words of a sentence. Therefore, Bi-LSTM
provides a way to include both previous and future context by
applying one LSTM to process sentence forward and another LSTM to
process sentence backward.

Given a sentence of $n$ words $\{ w_i \}_{i=1}^n$, we first transfer
the one-hot vector $ w_i $ into a dense vector $ x_i $ through an
embedding matrix $ W_e $ with the equation $ {x_i} = {W_e}{w_i}  $.
We use Bi-LSTM to get the annotations of words by processing
sentence from both directions. Bi-LSTM contains the backward $
\overleftarrow {LSTM} $ that reads the sentence from $ x_n $ to $
x_i $ and a forward $\overrightarrow {LSTM}$ which reads from $ x_1
$ to $ x_i $:
\begin{equation}
\begin{split}
{x_i} &= {W_e}{w_i},i \in \left[ {1,n} \right], \\
\mathop {{h_i}}\limits^ \to &= \overrightarrow {LSTM} ({x_i}),i \in \left[ {1,n} \right], \\
\mathop {{h_i}}\limits^ \leftarrow &= \overleftarrow {LSTM} ({x_i}),i \in \left[ {1,n} \right].
\end{split}
\end{equation}

At $i$ time step, we obtain the forward hidden state $\mathop
{{h_i}}\limits^ \to $ which stores previous information and the
backward hidden state $ \mathop {{h_i}}\limits^ \leftarrow $ which
stores future information. $ {h_i} = [{\mathop {{h_i}}\limits^
\to},{\mathop {{h_i}}\limits^ \leftarrow}] $ is a summary of the
sentence centered around $ w_i $.

\subsection{Attention and Multichannel}

For the NLP tasks such as text classification and sentiment
analysis, different words contribute unequally to the representation
of a sentence. The attention mechanism can be used to reflect the
importance weight of the input element so that the relevant element
contributes significantly to the merged output. Although the
attention mechanism is able to model dependencies flexibly, it is
still a crude process because of the loss of temporal order
information. We apply attention mechanisms to the hidden states of
Bi-LSTM and splice them into a matrix.

Taking the form of the matrix rather than a weighted sum of vectors
will keep the order information. Furthermore, by applying the scalar
attention and the vectorial attention, we could obtain several
matrices and take them as multichannel for inputs of CNN.

\subsubsection{Scalar Attention Mechanism}

We introduce the scalar attention to calculate the importance
weights of all input elements. $ M $ is the association matrix that
represents the association among words in texts. The element of the
$ i $-th row and the $ j $-th column of $ M $ represents the degree
of association between the $ i $-th word and the $ j $-th word. We
will set $ L $ channel mask matrices $ V $ if we need $ L $
channels. In the $ l $-th channel, $ M_{l_{i,j}} $ is calculated as
follows:
\begin{equation}
M_{l_{i,j}} = \tanh ( [{h_i} , {W_l} \cdot {h_j} ]  + b_l),
\end{equation}
The $ i $-th channel mask matrix is defined as follows:
\begin{equation}
V_{{l_{i,j}}} \sim B(1,{p_l}),i \in [1,n],j \in [1,n],
\end{equation}
That means each element of $ V_l $ obeys binomial distribution.
Given $ M_{l_{i,j}} $ and $ V_{{l_{i,j}}} $, the $ i $-th channel is computed as follows:
\begin{equation}
{A_l} = {M_l} \otimes {V_l},
\end{equation}
    \begin{equation}\label{sum_along}
{s_{lk}} = \sum\nolimits_x {{A_{l_{xk}}}},
\end{equation}
\begin{equation}
p_k = \\
\begin{cases}
-99999, \qquad  if\   x_k\  is\ from\ pad & \\
0,  \qquad \qquad       otherwise\ & \\
\end{cases}
\end{equation}
\begin{equation}
score_{lk} = p_k + s_{lk},
\end{equation}
\begin{equation}\label{scalar}
a_{lk} = \frac{{\exp (scor{e_{lk}})}}{{\sum\nolimits_{i = 1}^n {\exp (scor{e_{li}})} }},
\end{equation}

\begin{figure}
   \centering
   \includegraphics[width=0.5\linewidth]{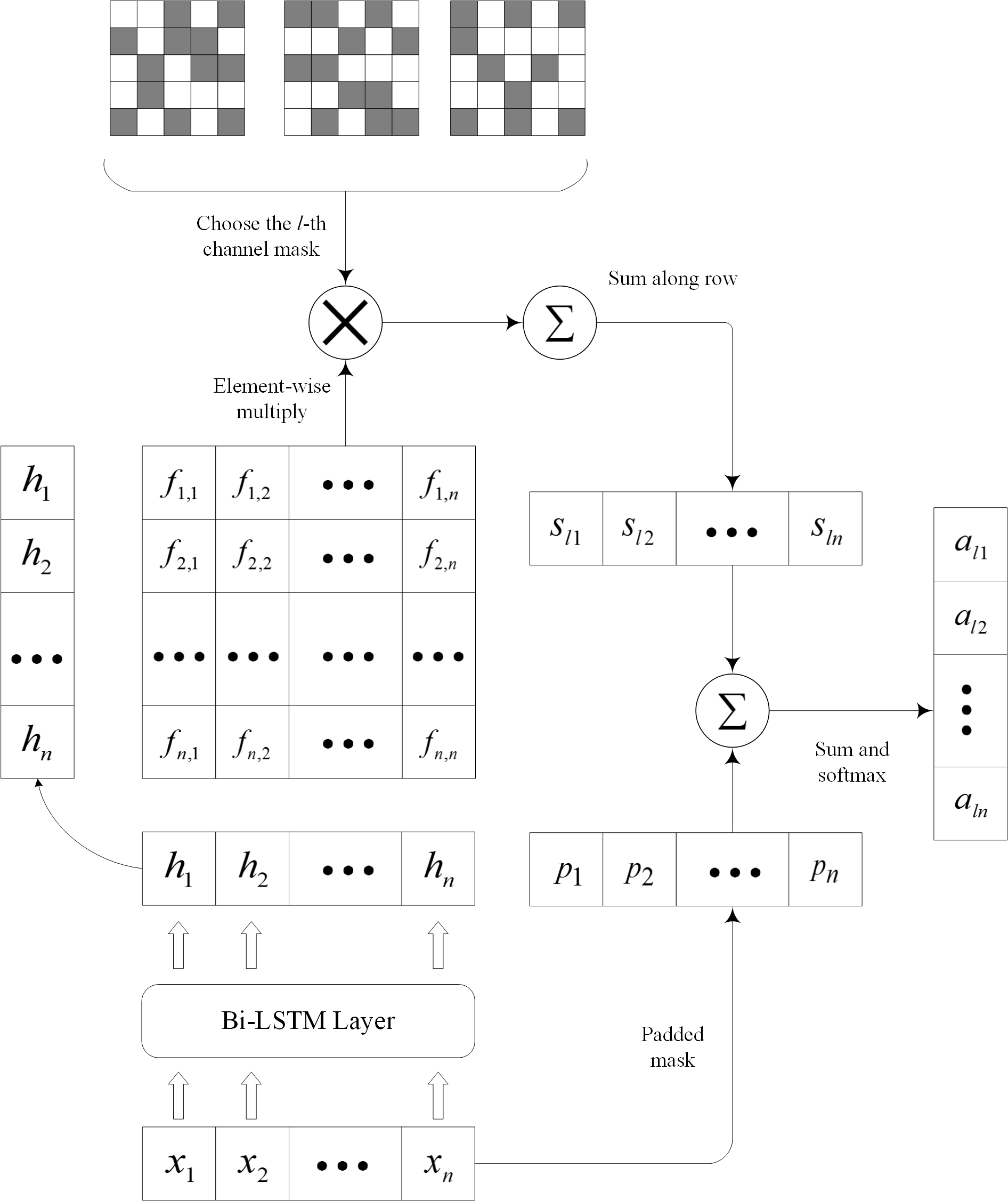}
   \caption{Scalar attention mechanism}
   \label{fig:scalar}
\end{figure}

\begin{equation}
{c_{li}} = {a_{li}}{\cdot}{h_i},
\end{equation}

\begin{equation}
{C_l} = \left[ {{c_{l1}},{c_{l2}},{c_{l3}}, \cdot  \cdot  \cdot  \cdot  \cdot  \cdot ,{c_{ln}}} \right].
\end{equation}
Here, $ {c_{li}}$ denotes the new representation of $ h_i $ in the $
l $-th channel and $ \otimes $ denotes element-wise product
operation. The $ pad $ symbol still carries little information after
it is encoded by Bi-LSTM. So, if word $ x_k $ is a $ pad $ symbol,
its scalar attention $ s_{lk} $ will be subtracted from 99999 before
softmax operation and so that $ a_{lk} $ will be close to 0 after
softmax. By concatenating all $C_{li}$, we obtain the $ l $-th
channel $ C_l $. The whole process of the scalar attention is shown
in Fig.~\ref{fig:scalar}.

\subsubsection{Vectorial Attention Mechanism}

Suppose a word or a sentence is encoded into an $ n $-dimensional
vector $ {\left( {{v_1},{v_2},{v_3},......,{v_n}} \right)^T} $ and
each dimension can be decoded into a specific meaning which
contributes differently to different tasks. For example, a sentence
consisting of two words ``$ Harvard $  $ University $" is encoded
into $ n $-dimension vector $ {\left( {{v_1},{v_2},......,{v_n}}
\right)^T} $, where $ v_1 $ can be decoded into ``$ America $" and $
v_2 $ can be decoded into ``$ Massachusetts $". Consequently, in the
classification task ``$ Does $  $ Harvard $  $ University $ $belong$
$ to$  $ America ? $" , $ v_1 $ is more informational than other
dimensions. But when given a QA task ``$ Which$  $ state$  $ is$  $
Harvard $  $ University $  $in ? $" , $ v_2 $ is more informational
than the others.

Based on the above assumptions, we propose the vectorial attention
mechanism to compute the vectorial importance weight of each
dimension in the input element:
\begin{equation}
\begin{split}
&\overrightarrow {scor{e_{li}}}  = {W_{l1}}^T\sigma \left( {{W_{l2}} \cdot {h_i} + {b_l}} \right), \\
&\mathop {{a_{li}}}\limits^ \to   = \frac{{\exp \left( {\overrightarrow {scor{e_{li}}} } \right)}}{{\sum\limits_i {\exp \left( {\overrightarrow {scor{e_{li}}} } \right)} }}, \\
&{c_{li}} = \mathop {{a_{li}}}\limits^ \to   \odot {h_{i}}, \\
&{C_l} = \left[ {{c_{l1}},{c_{l2}},{c_{l3}}, \cdot  \cdot  \cdot  \cdot  \cdot  \cdot ,{c_{ln}}} \right]. \\
\end{split}
\end{equation}
where $ c_{li} $ denotes the final representation of $ h_i $ in the $ l $-th channel. By concatenating all $c_{li}$ where $i \in \left[ {1,n} \right] $, we obtain the $ l $-th channel $ C_l $.

By combining the scalar attention and the vectorial attention,
multichannel is generated as follows:
\begin{equation}
\begin{split}
&{c_{li}} = {a_{li}}{\cdot}{(\mathop {{a_{li}}}\limits^ \to   \odot {h_{i}})}, \\
&{C_l} = \left[ {{c_{l1}},{c_{l2}},{c_{l3}}, \cdot  \cdot  \cdot  \cdot  \cdot  \cdot ,{c_{ln}}} \right]. \\
\end{split}
\end{equation}

Intuitively, the scalar attention mechanism introduces external
perturbation while the vectorial attention mechanism introduces
internal perturbation.

\subsubsection{Convolutional Neural Network}

Convolutional Neural Network (CNN) utilizes several sliding
convolution filters to extract local features. Assume we have one
channel that is represented as
\begin{equation}
C = \left[ {{c_1},{c_2},{c_3},...,{c_n}} \right].
\end{equation}
Here, $ C \in {\mathbb{R}^{n \times k}} $, $ n $ is the length of
the input element, and $k$ is the embedded dimension of each input
element. In a convolution operation, a filter $ {\bf{m}} \in
\mathbb{R}^{lk} $ is involved in applying to consecutive $ l $ words
to generate a new feature:
\begin{equation}
{x_i} = f\left( {{\bf{m}} \cdot {{\bf{c}}_{i:i + l - 1}} + b} \right),
\end{equation}
where $ {\bf{c}}_{i:i + l - 1} $ is the concatenation of $ {c_i},...,{c_{i+l-1}} $. $ f $ is a non-liner activation function such as $ relu $ and $ b \in \mathbb{R} $ is a bias term. After the filter $ {\bf{m}} $ slide across $ \left\{ {{{\bf{c}}_{1:l}},{{\bf{c}}_{2:l + 1}},...,{{\bf{c}}_{n - l + 1:n}}} \right\} $, we obtain a feature map:

\begin{equation}\label{key}
{\bf{x}} = \left[ {{x_1},{x_2},...,{x_{n - l + 1}}} \right].
\end{equation}

We apply max-pooling operation over the feature map $ {\bf{x}} $ and
take the maximum value $ \hat x   = \max \{ {\bf{x}}\}  $ as the
final feature extracted by the filter $  {\bf{m}} $. This pooling
scheme is to capture the most dominating feature for each filter.
CNN obtains multiple features by utilizing multiple filters with
varied sizes. These features form a vector $ {\bf{r}} = \left[
{{x_1},{x_2},...,{x_s}} \right] $ ($s$ is the number of filters)
which will be passed to a fully connected softmax layer to output
the probability distribution over labels

\begin{equation}
y = softmax \left( {W \cdot {\bf{r}} + b} \right).
\end{equation}

Given a training sample ($ \textbf{x}^i  $, $ y^i $) where $ {y^i}
\in \left\{ {1,2, \cdots ,c} \right\} $ is the true label of $
\textbf{x}^i $ and the estimated probability of our model is $
\tilde y_j^i \in [0,1] $ for each label $ j \in \left\{ {1,2, \cdots
,c} \right\} $, and the error is defined as:

\begin{equation}
L({{\bf{x}}^i},{y^i}) = -\sum\limits_{j = 1}^c {if\{ {y^i} = j\} } \log (\tilde y_j^i).
\end{equation}
Here, $ c $ denotes the number of possible labels of $ \textbf{x}^i
$ and $ if\{ \dot \}  $ is an indicator function such that: $ if\{
{y^i} = j \}=1  $ if $ {y^i} = j $, $ if\{ {y^i} = j \}=0  $
otherwise. We employ stochastic gradient descent (SGD) to update the
model parameters and adopt Adam optimizer.

\section{Experimental Study}
\label{sec:Experiment}

\subsection{Experiments Datasets}
We evaluate our model on several datasets. Summary statistics of the
datasets are shown in Table \ref{tab:data}.
\begin{table}
    \centering
    \begin{tabular}{lcccccc}
        \toprule
        Data  & $ c $ & $ l $ & $ N $ & $ V $ & $ V_{word} $ & $ Test $  \\
        \noalign{\smallskip}\hline\noalign{\smallskip}
        MR & 2 & 20  & 10662 & 18765 & 16448 & CV \\
        Subj & 2 & 23  & 10000 & 21323 & 17913 & CV \\
        MPQA & 2 & 3  & 10606 & 6246 & 6083 & CV \\
        SST-1 & 5 & 18  & 11855 & 17836 & 16262 & 2210 \\
        SST-2 & 2 & 19  & 9613 & 16185 & 14838 & 1821 \\
        \bottomrule
    \end{tabular}
    \caption{Summary statistics of the datasets. $ c $: Number of classes. $ l $: Average length of sentences. $ N $: Size of datasets. $ V $: Vocabulary size. $ V_{word} $: Number of words present in the set of pre-trained word vectors, respectively. $ Test $: Size of test sets. $ CV $(cross validation): No standard train/test split and thus 10-fold CV was used. }
    \label{tab:data}
\end{table}

\begin{itemize}
    \item  \textbf{MR:} Short movie review dataset with one sentence per review. Each review was labeled with their overall sentiment polarity (positive or negative).

    \item \textbf{Subj:} Subjectivity dataset containing sentences labeled
    with respect to their subjectivity status (subjective or objective).

    \item \textbf{SST-1:} Stanford Sentiment Treebank—an
    extension of MR but with train/dev/test splits
    provided and fine-grained labels (very positive, positive, neutral, negative, very negative).

    \item \textbf{SST-2:} Same as SST-1 but with neutral reviews removed and binary labels

    \item \textbf{MPQA:} Opinion polarity detection subtask of the MPQA dataset.
\end{itemize}

\begin{table*}
    \centering
    \setlength{\tabcolsep}{7mm}{
        \begin{tabular}{lccccc}
            \toprule

            Model & MR & Subj & MPQA & SST-1 & SST-2 \\
            \noalign{\smallskip}\hline\noalign{\smallskip}
            Sent-Paser\cite{dong2015statistical} & 79.5 & -  & 86.3  & - & - \\

            NBSVM\cite{wang2012baselines} & 79.4 & 93.2  & 86.3 & - & - \\

            MNB\cite{wang2012baselines} & 79.0 & 93.6  & 86.3 & - & - \\

            F-Dropout\cite{wang2013fast} & 79.1 & 93.6  & 86.3 & - & - \\

            G-Dropout\cite{wang2013fast} & 79.0 & 93.4  & 86.1 & - & - \\

            \noalign{\smallskip}\hline\noalign{\smallskip}

            Paragraph-Vec\cite{le2014distributed} & - & -  & - & 48.7 & 87.8 \\

            RAE\cite{socher2011semi} & 77.7 & -  & - & 43.2 & 82.4  \\

            MV-RNN\cite{socher2012semantic} & 79.0 & -  & - & 44.4 & 82.9  \\

            RNTN\cite{socher2013recursive} & - & - & -& 45.7  & 85.4   \\

            DCNN\cite{kalchbrenner2014convolutional} & - & - & -& \underline{48.5}  & 86.8   \\

            Fully Connected\cite{limsopatham2016modelling} & 81.59 & -  & - & - & -  \\

            CNN-non-static\cite{kim2014convolutional} & 81.5 & 93.4  & 89.5 & 48.0 & 87.2  \\

            CNN-multichannel \cite{kim2014convolutional} & 81.1 & 93.2 & 89.4 & 47.4 & \underline{88.1}  \\

            WkA+25\%fiexible\cite{lakshmana2016learning} & 80.02 & 92.68 & - & 46.11 & 84.29 \\

            Fully Connected \cite{limsopatham2016modelling} & 81.59 & - & - & - & - \\

            L-MCNN \cite{guo2018integrated} & 82.4 & - & - & - & - \\

            Hclustering avg \cite{kim2018cnn} & 80.20 & - & - & - & - \\

            Kmeans centroid  \cite{kim2018cnn} & 80.21 & - & - & - & - \\

            \noalign{\smallskip}\hline\noalign{\smallskip}

            AMCNN-1 & 82.17 & 92.96 & 89.61 & 47.02  & 86.43   \\
            AMCNN-3 & \underline{82.57} & \underline{93.75} & \underline{89.75} & 47.58  & 86.85  \\
            AMCNN-5 & 82.41 & 93.43 & 89.34 & 47.41  & 86.56   \\
            AMCNN-7 & 82.23 & 93.36 & 89.46 & 47.16  & 86.29   \\
            AMCNN-rv & 82.34 & 93.52 & 89.55 & 47.37  & 86.69   \\
            \bottomrule
    \end{tabular}}
    \caption{\small{Results of our AMCNN models against other models. We use underline to highlight wins. }}
    \label{tab:result}
\end{table*}

\subsection{Experiments Settings}

\begin{itemize}
    \item \textbf{Padding:} We first use $ len $ to denote the maximum length of the sentence in the training set. As the convolution layer requires input of fixed length, we pad each sentence that has a length less than $ len $ with $ UNK $ symbol which indicates the unknown word in front of the sentence. Sentences in the test dataset that are shorter than $ len $ are padded in the same way, but for sentences that have a length longer that $ len $, we just cut words at the end of these sentences to ensure all sentences have a length $ len $.

    \item \textbf{Initialization:} We use publicly available $word2vec$ vectors to initialize the words in the dataset. $ word2vec $ vectors are pre-trained on 100 billion words from Google News through an unsupervised neural language model.
    For words that are not present in the set of pre-trained words ore rarely appear in data sets, we initialize each dimension from $ U\left[ { - 0.25,0.25} \right] $ to ensure all word vectors have the same variance. Word vectors are fine-tuning along with other parameters during the training process.

    \item \textbf{Hyper-parameters:} The feature representation of Bi-LSTM is controlled by the size of hidden states. We investigate our model with various hidden sizes and set the hidden size of unidirectional LSTM to be 100. We also investigate the impact of the size of the channels on our model. When the size of the channels is set to be 1, our model is a single channel network. When increasing the size of the channels, our model obtains a more semantic representation of the text. Too many channels increase parameters and thus it can cause overfitting. Eclectically, we simply set the size of the channels to be 3. Convolutional filter decides the n-gram feature which directly influences the classification performance. We
    set the filter size based on different datasets and simply set the filter map to be 100. More details of hyper-parameters are shown on Table \ref{tab:set}.

\begin{table}
    \centering
    \begin{tabular}{lcc}
        \toprule
        Hyperparameter  &   Ranges  &  Adopt\\
        \noalign{\smallskip}\hline\noalign{\smallskip}
        $ Hidden\ size $ & \{16, 32, 50, 64, 100, 128, $ \cdots $\} & 100 \\
        $ L2 $ & \{0.0001, 0.0005, 0.001, 0.003, $ \cdots $\} & 0.0005\\
        $ Channel $ & \{1, 2, 3, 4, 5, 6, $ \cdots $\} &  3 \\
        $ Filter\ size $ & \{(2,3,4), (3,4,5), (4,5,6), $ \cdots $\} & -\\
        $ Filter\ map $ & \{10, 30, 50, 100, 150, $ \cdots $\} & 100\\
        \bottomrule
    \end{tabular}
    \caption{Hyper-parameters setting. $ Hidden\ size $: The dimension of unidirectional LSTM. $ L2 $: $ L2 $ regularization term. $ Channel $: The number of channels. $ Filter\ size $: The size of convolutional filters. $ Filter\ map $: The number of convolutional filter maps.}
    \label{tab:set}       
\end{table}

    \item  \textbf{Other settings:}  We only use one Bi-LSTM layer and one convolutional layer. Dropout is applied on the word embedding layer, the CNN input layer, and the penultimate layer. Weight vectors are constrained by $ L2 $ regularization  and the model is trained to minimize the cross-entropy loss of true labels and the predicted labels.

\end{itemize}

\begin{figure}
   \centering
   \includegraphics[width=0.9\linewidth]{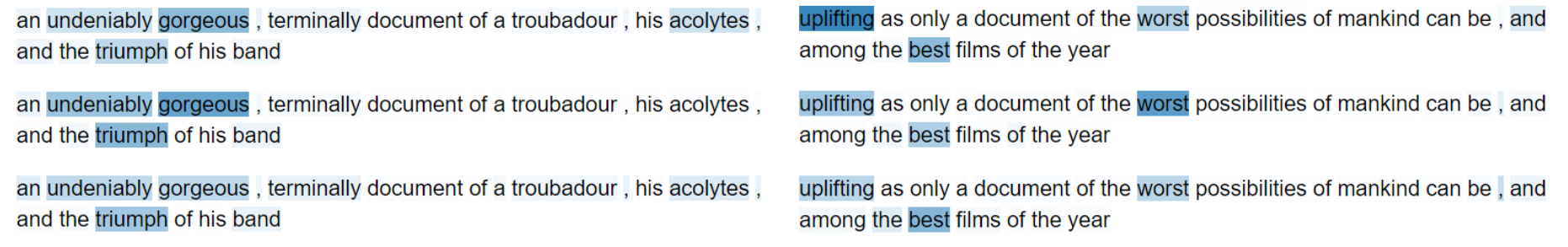}
   \caption{Visualization of the scalar attention weights learned by different channels}
   \label{img}
\end{figure}

\subsection{Baselines}
We compare our model with several baseline methods which can be divided into the following categories:

\paragraph{$ \bf{Traditional\ Machine\ Learning} $} A statistical parsing framework was studied for sentence-level sentiment classification\cite{dong2015statistical}. Simple Naive Bayes (NB)
and Support Vector Machine (SVM) variants outperformed most
published results on sentiment analysis
datasets\cite{wang2012baselines}. It was shown in
\cite{wang2013fast} how to do fast dropout training by sampling from
or integrating a Gaussian approximation. These measures were
justified by the central limit theorem and empirical evidence, and
they resulted in an order of magnitude speedup and more stability.

\paragraph{$ \bf{Deep\ Learning} $} Word2vec \cite{le2014distributed} was extended with a new method called Paragraph-Vec, which is an unsupervised algorithm that learns fixed-length feature representations from variable-length pieces of texts, such as sentences, paragraphs, and documents.
Various recursive networks were extended
\cite{socher2012semantic,socher2011semi,socher2013recursive}.
Generic and target domain embeddings were incorporated to
CNN\cite{kalchbrenner2014convolutional}. A series of experiments
with CNNs was trained on top of pre-trained word vectors for
sentence-level classification tasks \cite{kim2014convolutional}.
Desirable properties such as semantic coherence, attention mechanism
and kernel reusability in CNN were empirically studied for learning
sentence-level tasks \cite{lakshmana2016learning}. Both word
embeddings created from generic and target domain corpora were
utilized when it's difficult to find a domain corpus
\cite{limsopatham2016modelling}. A hybrid L-MCNN model was proposed
to represent the semantics of sentences \cite{guo2018integrated}.
\begin{figure*}
    \centering
      \subfigure[]{\label{hidden}
       \includegraphics[width=0.45\linewidth]{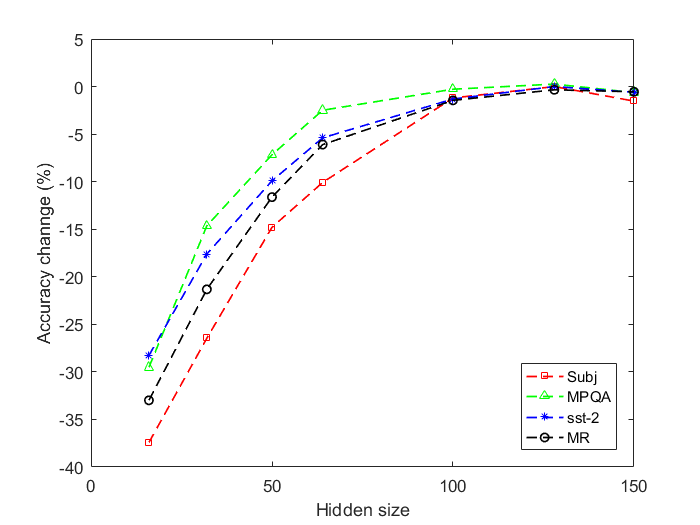}}
    \label{1a}\hfill
      \subfigure[]{\label{channel}
        \includegraphics[width=0.45\linewidth]{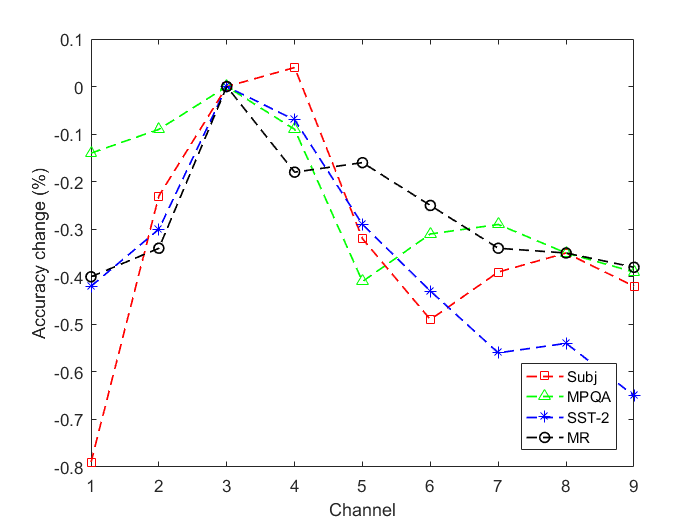}}
    \label{1b}\\
      \subfigure[]{\label{filter}
        \includegraphics[width=0.45\linewidth]{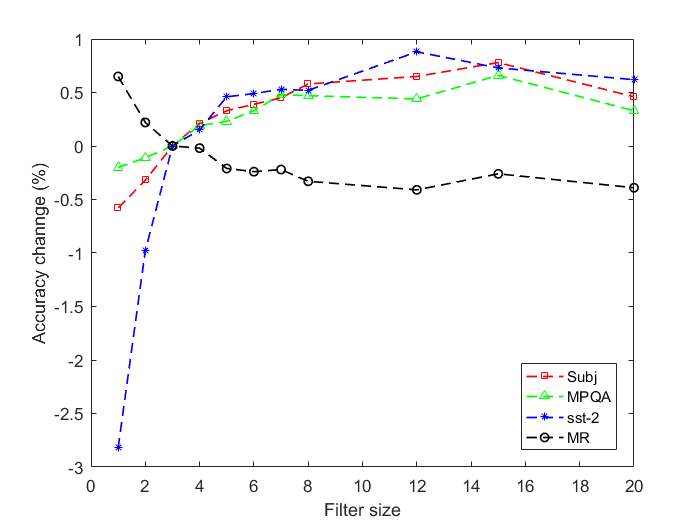}}
    \label{1c}\hfill
      \subfigure[]{\label{map}
        \includegraphics[width=0.45\linewidth]{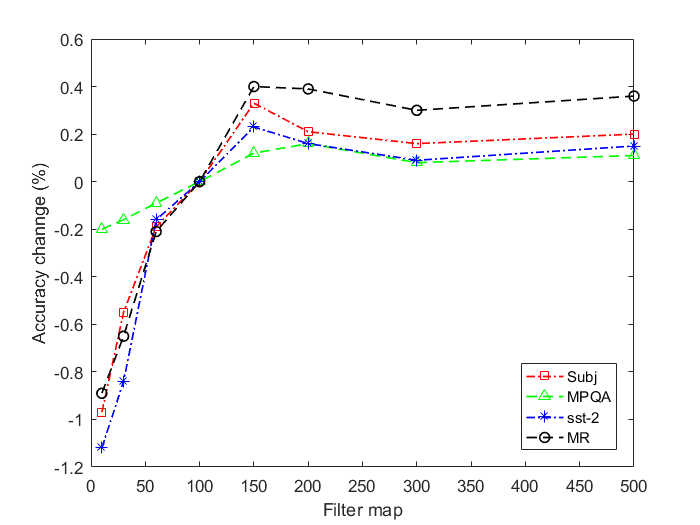}}
      \caption{Effect of hyper-parameters: (a) hidden size, (b) channel,  (c) filter size , and (d) the filter map.}
\end{figure*}

\subsection{Results and Analysis}

Table~\ref{tab:result} shows results of our model on five datasets against other methods. We refer to our model as AMCNN-$ \{1,3,5,7\} $. Here AMCNN-$ x $ stands for AMCNN with $ x $ channels which means there are $ x $ different representations of the text. As we can see, our model exceeds in 3 out of 5 tasks. For MR/Subj/MPQA, AMCNN-$ 3 $ outperforms other baselines and we can get a rough observation that AMCNN-3 performs better than AMCNN-\{5,7\}, and they all perform better than AMCNN-1, which is a single channel model. This phenomenon indicates that multichannel representation is effective,  but continuing to increase the size of the channels does not improve our model all the time. We conjecture that it would be better to choose $ x $ according to the number of informative  words in the sentence. Take the following sentences for example:
\begin{enumerate}
    \item $ An $  $ undeniably $  $ gorgeous, $  $ terminally $  $ document $  $ of $  $ a $  $ troubadour, $  $ his $  $ acolytes, $  $ and $  $ the $  $ triumph $  $ of $  $ his $  $ band $.

    \item $ Uplifting $  $ as $ $  only $ $ a $ $ document $ $ of $ $ the $ $ worst $ $ possibilities $ $of$ $mankind$ $can$ $be$, $and$ $among$ $the$ $ best $ $films$ $of$ $the$ $year$.
\end{enumerate}

Fig.~\ref{img} shows the visualization of scalar attention distribution of the above sentences.

The second sentence could not be labeled positive or negative without a  doubt if we focus on a single informative  word (``$uplifting$", ``$worst$" or ``$best$") alone. Only if these informative  words were all emphasized can this sentence be truly understood. ``$ Uplifting $" received more attention weight than other words in the first channel. ``$ worst $" received more attention weight in the second than the third channel and ``$ best $" received more attention weight in the third than the second channel. If the second channel is set to be an independent model, this sentence might be classified incorrectly. But AMCNN-3 will still label this sentence as positive.
Multichannel essentially provides a way to represent a sentence from different views and provides diversification.

We also investigate the impact of the vectorial attention on AMCNN and find that it improves performance. AMCNN-rv denotes AMCNN-3 without applying the  vectorial attention mechanism. We owe the validity of the  vectorial attention mechanism to its selectivity of features that can better describe the word in the specific given tasks. Actually, the  vectorial attention mechanism introduces the perturbation of hidden states and makes our model more robust. Another advantage of the  vectorial attention mechanism is that it assigns different learning speeds to each dimension of the hidden state indirectly so that informative dimension could be tuned at a bigger pace than dimension of less information.

%

\subsection{Parameter Sensitivity}

We further evaluate how the parameters of AMCNN impact its performance on the text classification task. In this experiment, we evaluate the effect of change of $ Hidden\ size $, $ Channel $, $ Filter\ size $, and $ Filter\ map $ on AMCNN performance with other parameters remaining the same.

\begin{itemize}
    \item  \textbf{Impact of \textit{Hidden size}: }  Fig.\ref{hidden} shows the impact of \textit{Hidden size} on classification accuracy. It can be observed that the classification accuracy of the model increases with the increasing of hidden size. When the hidden size is set to be 128, the accuracy curve of the model tends to be flat or even begins to decline.
    So, the hidden size of Bi-LSTM affects the encoding of the document. If the \textit{Hidden size} is too small, it will lead to underfitting. If the \textit{Hidden size} is too large, it will lead to overfitting.

    \item  \textbf{Impact of \textit{Channel}: }   Fig.\ref{channel} shows the impact of \textit{Channel} on classification accuracy. We observe that the performance first rises and then tends to decline. When channel size is set to be 3, the model (AMCNN-3) performs best on MPQA/SST-2/MR datasets. The model (AMCNN-4) performs best on Subj dataset when channel size is set to be 4. This result shows that multichannel representations of texts help our model improve its performance. However, as the number of channels increases, the parameters of the model also increase, which may lead to overfitting.

    \item  \textbf{Impact of \textit{Filter size}: }  Fig.\ref{filter} shows the impact of \textit{Filter size} on classification accuracy. It can be observed that the optimal filter size settings of each dataset are different, and the accuracy curve of the MR dataset is opposite to the accuracy curve of other datasets. When \textit{Filter size} is between [10, 14], the model achieves high accuracy on MPQA/Subj/SST-2 datasets. But this performance improvement is not significant compared to the accuracy when \textit{Filter size} is 2.
    In order to reduce the size of the parameters, \textit{Filter size} of the model is set between [4, 8] in the experiment.

    \item  \textbf{Impact of \textit{Filter map}: }  Fig.\ref{map} shows the impact of \textit{Filter map} on classification accuracy. We can observe that the performance rises rapidly first and then tends to be flat.
    The number of \textit{Filter map} determines the number of feature maps generated after the convolution operation. Each feature map represents a certain feature of the text. The more the number of feature maps, the more features that the convolution operation can extract, and the accuracy of the model can be higher.
    But the number of features of the text is finite, and the increase in the number of \textit{Filter map} will also increase the size of trainable parameters,  which may lead to overfitting.
\end{itemize}

\section{Conclusion and Future Work}
\label{sec:Conclusion}

In this paper, we propose Attention-based Multichannel Convolutional
Neural Network (AMCNN) for text classification. Our model applies
Bi-LSTM to capture contextual information and obtains multichannel
representations by using the scalar attention and the vectorial
attention. AMCNN is able to extract n-gram features with context
attached and introduce perturbation to provide robustness. The
experimental results demonstrate that the AMCNN model achieves
superior performance on the text classification task. Visualization
of attention distribution illustrates that multichannel is effective
in capturing informative words of different perspectives.

In the future, we will focus on applying a generative model to
obtain multichannel representations of texts. Data augmentation of
texts could be realized in this way and texts could be represented
in rich semantics. CNN requires fixed-length inputs and performs
unnecessary convolution operations when dealing with NLP tasks. So,
we will also explore CNN architecture that can process the text with
variable length.

\bibliographystyle{unsrt}
\bibliography{bibtex}

\begin{thebibliography}{10}

\bibitem{Jiang2018LatentTT}
B.~Jiang, Z.~Li, H.~Chen, and A.~G. Cohn.
\newblock Latent topic text representation learning on statistical manifolds.
\newblock {\em IEEE Transactions on Neural Networks and Learning Systems},
  29:5643--5654, 2018.

\bibitem{vapnik1998statistical}
Vladimir~Naumovich Vapnik.
\newblock An overview of statistical learning theory.
\newblock {\em IEEE Transactions on Neural Networks}, 10(5):988--999, 1999.

\bibitem{chen-2009-pcvm}
H.~{Chen}, P.~{Tino}, and X.~{Yao}.
\newblock Probabilistic classification vector machines.
\newblock {\em IEEE Transactions on Neural Networks}, 20(6):901--914, 2009.

\bibitem{chen-2014-epcvm}
H.~{Chen}, P.~{Tiňo}, and X.~{Yao}.
\newblock Efficient probabilistic classification vector machine with
  incremental basis function selection.
\newblock {\em IEEE Transactions on Neural Networks and Learning Systems},
  25(2):356--369, 2014.

\bibitem{lyu2019multiclass}
S.~Lyu, X.~Tian, Y.~Li, B.~Jiang, and H.~Chen.
\newblock Multiclass probabilistic classification vector machine.
\newblock {\em IEEE Transactions on Neural Networks and Learning Systems},
  2019.

\bibitem{jiang2017scalable}
B.~Jiang, H.~Chen, B.~Yuan, and X.~Yao.
\newblock Scalable graph-based semi-supervised learning through sparse bayesian
  model.
\newblock {\em IEEE Transactions on Knowledge and Data Engineering},
  29(12):2758--2771, 2017.

\bibitem{bengio2003neural}
Yoshua Bengio, R{\'e}jean Ducharme, Pascal Vincent, and Christian Jauvin.
\newblock A neural probabilistic language model.
\newblock {\em Journal of machine learning research}, 3(Feb):1137--1155, 2003.

\bibitem{le2014distributed}
Quoc Le and Tomas Mikolov.
\newblock Distributed representations of sentences and documents.
\newblock In {\em International Conference on Machine Learning}, pages
  1188--1196, 2014.

\bibitem{luong2015effective}
Minh-Thang Luong, Hieu Pham, and Christopher~D Manning.
\newblock Effective approaches to attention-based neural machine translation.
\newblock {\em arXiv preprint arXiv:1508.04025}, 2015.

\bibitem{zhang2015character}
Xiang Zhang, Junbo Zhao, and Yann LeCun.
\newblock Character-level convolutional networks for text classification.
\newblock In {\em Advances in neural information processing systems}, pages
  649--657, 2015.

\bibitem{xu-etal-2016-improve}
J.~Xu, J.~Liu, L.~Zhang, Z.~Li, and H.~Chen.
\newblock Improve {C}hinese word embeddings by exploiting internal structure.
\newblock In {\em Proceedings of NAACL}, pages 1041--1050, 2016.

\bibitem{tang2015document}
Duyu Tang, Bing Qin, and Ting Liu.
\newblock Document modeling with gated recurrent neural network for sentiment
  classification.
\newblock In {\em Proceedings of the 2015 conference on empirical methods in
  natural language processing}, pages 1422--1432, 2015.

\bibitem{hochreiter1997long}
Sepp Hochreiter and J{\"u}rgen Schmidhuber.
\newblock Long short-term memory.
\newblock {\em Neural computation}, 9(8):1735--1780, 1997.

\bibitem{zhao2020condition}
X.~Zhao, X.~Feng, H.~Zhong, J.~Yao, and H.~Chen.
\newblock Condition aware and revise transformer for question answering.
\newblock In {\em Proceedings of The Web Conference 2020}, pages 2377--2387,
  2020.

\bibitem{cho2014learning}
Kyunghyun Cho, Bart Van~Merri{\"e}nboer, Caglar Gulcehre, Dzmitry Bahdanau,
  Fethi Bougares, Holger Schwenk, and Yoshua Bengio.
\newblock Learning phrase representations using rnn encoder-decoder for
  statistical machine translation.
\newblock {\em arXiv preprint arXiv:1406.1078}, 2014.

\bibitem{lei2017training}
Tao Lei, Yu~Zhang, and Yoav Artzi.
\newblock Training rnns as fast as cnns.
\newblock {\em arXiv preprint arXiv:1709.02755}, 2017.

\bibitem{kalchbrenner2014convolutional}
Nal Kalchbrenner, Edward Grefenstette, and Phil Blunsom.
\newblock A convolutional neural network for modelling sentences.
\newblock {\em arXiv preprint arXiv:1404.2188}, 2014.

\bibitem{yih2014semantic}
Wen-tau Yih, Xiaodong He, and Christopher Meek.
\newblock Semantic parsing for single-relation question answering.
\newblock In {\em Proceedings of the 52nd Annual Meeting of the Association for
  Computational Linguistics (Volume 2: Short Papers)}, volume~2, pages
  643--648, 2014.

\bibitem{kim2014convolutional}
Yoon Kim.
\newblock Convolutional neural networks for sentence classification.
\newblock {\em arXiv preprint arXiv:1408.5882}, 2014.

\bibitem{lai2015recurrent}
Siwei Lai, Liheng Xu, Kang Liu, and Jun Zhao.
\newblock Recurrent convolutional neural networks for text classification.
\newblock In {\em AAAI}, volume 333, pages 2267--2273, 2015.

\bibitem{wang2017hybrid}
Chenglong Wang, Feijun Jiang, and Hongxia Yang.
\newblock A hybrid framework for text modeling with convolutional rnn.
\newblock In {\em Proceedings of the 23rd ACM SIGKDD International Conference
  on Knowledge Discovery and Data Mining}, pages 2061--2069. ACM, 2017.

\bibitem{mikolov2013distributed}
Tomas Mikolov, Ilya Sutskever, Kai Chen, Greg~S Corrado, and Jeff Dean.
\newblock Distributed representations of words and phrases and their
  compositionality.
\newblock In {\em Advances in neural information processing systems}, pages
  3111--3119, 2013.

\bibitem{yang2016hierarchical}
Zichao Yang, Diyi Yang, Chris Dyer, Xiaodong He, Alex Smola, and Eduard Hovy.
\newblock Hierarchical attention networks for document classification.
\newblock In {\em Proceedings of the 2016 Conference of the North American
  Chapter of the Association for Computational Linguistics: Human Language
  Technologies}, pages 1480--1489, 2016.

\bibitem{joulin2016bag}
Armand Joulin, Edouard Grave, Piotr Bojanowski, and Tomas Mikolov.
\newblock Bag of tricks for efficient text classification.
\newblock {\em arXiv preprint arXiv:1607.01759}, 2016.

\bibitem{conneau2016very}
Alexis Conneau, Holger Schwenk, Lo{\"\i}c Barrault, and Yann Lecun.
\newblock Very deep convolutional networks for text classification.
\newblock {\em arXiv preprint arXiv:1606.01781}, 2016.

\bibitem{mou2015discriminative}
Lili Mou, Hao Peng, Ge~Li, Yan Xu, Lu~Zhang, and Zhi Jin.
\newblock Discriminative neural sentence modeling by tree-based convolution.
\newblock {\em arXiv preprint arXiv:1504.01106}, 2015.

\bibitem{yin2016multichannel}
Wenpeng Yin and Hinrich Sch{\"u}tze.
\newblock Multichannel variable-size convolution for sentence classification.
\newblock {\em arXiv preprint arXiv:1603.04513}, 2016.

\bibitem{li2018symbolic}
Y.~Li, B.~Jiang, H.~Chen, and X.~Yao.
\newblock Symbolic sequence classification in the fractal space.
\newblock {\em IEEE Transactions on Emerging Topics in Computational
  Intelligence}, 2018.

\bibitem{gong2018sequential}
Z.~Gong and H.~Chen.
\newblock Sequential data classification by dynamic state warping.
\newblock {\em Knowledge and Information Systems}, 57(3):545--570, 2018.

\bibitem{chen2014cognitive}
H.~Chen, P.~Ti{\v{n}}o, and X.~Yao.
\newblock Cognitive fault diagnosis in tennessee eastman process using learning
  in the model space.
\newblock {\em Computers \& chemical engineering}, 67:33--42, 2014.

\bibitem{ChenTRY14}
H.~Chen, P.~Ti{\~{n}}o, A.~Rodan, and X.~Yao.
\newblock Learning in the model space for cognitive fault diagnosis.
\newblock {\em {IEEE} Transactions Neural Networks Learning System},
  25(1):124--136, 2014.

\bibitem{chen2013model}
H.~Chen, F.~Tang, P.~Tino, and X.~Yao.
\newblock Model-based kernel for efficient time series analysis.
\newblock In {\em Proceedings of the 19th ACM SIGKDD international conference
  on Knowledge discovery and data mining}, pages 392--400, 2013.

\bibitem{gong2016model}
Z.~Gong and H.~Chen.
\newblock Model-based oversampling for imbalanced sequence classification.
\newblock In {\em Proceedings of the 25th ACM International on Conference on
  Information and Knowledge Management}, pages 1009--1018, 2016.

\bibitem{chen2015model}
H.~Chen, F.~Tang, P.~Tino, A.~G. Cohn, and X.~Yao.
\newblock Model metric co-learning for time series classification.
\newblock In {\em Twenty-Fourth International Joint Conference on Artificial
  Intelligence}, 2015.

\bibitem{gong2018multiobjective}
Z.~Gong, H.~Chen, B.~Yuan, and X.~Yao.
\newblock Multiobjective learning in the model space for time series
  classification.
\newblock {\em IEEE Transactions on Cybernetics}, 49(3):918--932, 2018.

\bibitem{chung2014empirical}
Junyoung Chung, Caglar Gulcehre, KyungHyun Cho, and Yoshua Bengio.
\newblock Empirical evaluation of gated recurrent neural networks on sequence
  modeling.
\newblock {\em arXiv preprint arXiv:1412.3555}, 2014.

\bibitem{bahdanau2014neural}
Dzmitry Bahdanau, Kyunghyun Cho, and Yoshua Bengio.
\newblock Neural machine translation by jointly learning to align and
  translate.
\newblock {\em arXiv preprint arXiv:1409.0473}, 2014.

\bibitem{vaswani2017attention}
Ashish Vaswani, Noam Shazeer, Niki Parmar, Jakob Uszkoreit, Llion Jones,
  Aidan~N Gomez, {\L}ukasz Kaiser, and Illia Polosukhin.
\newblock Attention is all you need.
\newblock In {\em Advances in Neural Information Processing Systems}, pages
  5998--6008, 2017.

\bibitem{bengio1994learning}
Yoshua Bengio, Patrice Simard, and Paolo Frasconi.
\newblock Learning long-term dependencies with gradient descent is difficult.
\newblock {\em IEEE transactions on neural networks}, 5(2):157--166, 1994.

\bibitem{dong2015statistical}
Li~Dong, Furu Wei, Shujie Liu, Ming Zhou, and Ke~Xu.
\newblock A statistical parsing framework for sentiment classification.
\newblock {\em Computational Linguistics}, 41(2):293--336, 2015.

\bibitem{wang2012baselines}
Sida Wang and Christopher~D Manning.
\newblock Baselines and bigrams: Simple, good sentiment and topic
  classification.
\newblock In {\em Proceedings of the 50th Annual Meeting of the Association for
  Computational Linguistics: Short Papers-Volume 2}, pages 90--94. Association
  for Computational Linguistics, 2012.

\bibitem{wang2013fast}
Sida Wang and Christopher Manning.
\newblock Fast dropout training.
\newblock In {\em international conference on machine learning}, pages
  118--126, 2013.

\bibitem{socher2011semi}
Richard Socher, Jeffrey Pennington, Eric~H Huang, Andrew~Y Ng, and
  Christopher~D Manning.
\newblock Semi-supervised recursive autoencoders for predicting sentiment
  distributions.
\newblock In {\em Proceedings of the conference on empirical methods in natural
  language processing}, pages 151--161. Association for Computational
  Linguistics, 2011.

\bibitem{socher2012semantic}
Richard Socher, Brody Huval, Christopher~D Manning, and Andrew~Y Ng.
\newblock Semantic compositionality through recursive matrix-vector spaces.
\newblock In {\em Proceedings of the 2012 joint conference on empirical methods
  in natural language processing and computational natural language learning},
  pages 1201--1211. Association for Computational Linguistics, 2012.

\bibitem{socher2013recursive}
Richard Socher, Alex Perelygin, Jean Wu, Jason Chuang, Christopher~D Manning,
  Andrew Ng, and Christopher Potts.
\newblock Recursive deep models for semantic compositionality over a sentiment
  treebank.
\newblock In {\em Proceedings of the 2013 conference on empirical methods in
  natural language processing}, pages 1631--1642, 2013.

\bibitem{limsopatham2016modelling}
Nut Limsopatham and Nigel Collier.
\newblock Modelling the combination of generic and target domain embeddings in
  a convolutional neural network for sentence classification.
\newblock {\em ACL 2016}, page 136, 2016.

\bibitem{lakshmana2016learning}
Madhusudan Lakshmana, Sundararajan Sellamanickam, Shirish Shevade, and Keerthi
  Selvaraj.
\newblock Learning semantically coherent and reusable kernels in convolution
  neural nets for sentence classification.
\newblock {\em arXiv preprint arXiv:1608.00466}, 2016.

\bibitem{guo2018integrated}
Yanbu Guo, Weihua Li, Chen Jin, Yunhao Duan, and Shuang Wu.
\newblock An integrated neural model for sentence classification.
\newblock In {\em 2018 Chinese Control And Decision Conference (CCDC)}, pages
  6268--6273. IEEE, 2018.

\bibitem{kim2018cnn}
Hwa-Yeon Kim, Jinsu Lee, Na~Young Yeo, Marcella Astrid, Seung-Ik Lee, and
  Young-Kil Kim.
\newblock Cnn based sentence classification with semantic features using word
  clustering.
\newblock In {\em 2018 International Conference on Information and
  Communication Technology Convergence (ICTC)}, pages 484--488. IEEE, 2018.

\end{thebibliography}

\end{document}